\documentclass[twocolumn]{article}

\usepackage[width=17cm,height=22cm]{geometry}
\usepackage[utf8]{inputenc}
\usepackage{fancyvrb}
\usepackage{authblk}
\usepackage{hyperref}
\usepackage[cmex10]{amsmath}
\usepackage{graphicx}
\usepackage{float}
\usepackage[caption = false]{subfig}

\graphicspath{{./}{./images/}}

\title{Using SVDD in SimpleMKL for 3D-Shapes Filtering}
\author{Gaëlle Loosli and Hattoibe Aboubacar}
\affil{Clermont-Université, Université Blaise Pascal, LIMOS, BP 10448, F-63000 Clermont-Ferrand \\ CNRS, UMR 6158, LIMOS, F-63173 Aubière}
\date{}
\begin{document}
\maketitle
\begin{abstract}
  This paper proposes the adaptation of Support Vector Data Description (SVDD) to the multiple kernel case (MK-SVDD), based on SimpleMKL. It also introduces a variant called Slim-MK-SVDD that is able to produce  a tighter frontier around the data. For the sake of comparison, the equivalent methods are also developed for One-Class SVM, known to be very similar to SVDD for certain shapes of kernels. 
  Those algorithms are illustrated in the context of 3D-shapes filtering and outliers detection. For the 3D-shapes problem, the objective is to be able to select a sub-category of 3D-shapes, each sub-category being learned with our algorithm in order to create a filter. For outliers detection, we apply the proposed algorithms for unsupervised outliers detection as well as for the supervised case.  
\end{abstract}

\medskip

\noindent\textbf{Mots-clef}: SVDD, SimpleMKL, One-Class SVM, 3D-shapes Filtering, Outliers detection.
\section{Introduction}
\label{sec:intro}
Multiple Kernel Learning (MKL) have encountered a great success since it was introduced \cite{lanckriet2004learning,sonnenburg2006large,bach2004multiple}. In \cite{rakotomamonjy2008simplemkl}, the SimpleMKL algorithm has the nice particularity that different solvers for different tasks can be plugged in it (given that it is possible to compute the cost function's gradient). Here we derive SimpleMKL for Support Vector Data Description (SVDD). Observing (see section \ref{sec:2Dtoy}) that the multiple kernel setting tends to favor large bandwidth kernels (in the case of Gaussian kernels) or low degree kernels (in the case of polynomial kernels), we also explore a different penalization criteria in order to produce tighter boundaries around the target class. 
Then we compare the resulting algorithms to SVDD alone, One-Class SVM and SimpleMKL with OneClass SVM.

\section{SVDD in SimpleMKL}
\label{sec:SVDDinSimpleMKL}
SVDD was introduced in \cite{tax2004support} to deal with one class classification or outliers detection. We first recall its formulation, before deriving it for the MKL case. Then we propose the adaptation to the SimpleMKL algorithm.

\subsection{SVDD}
We first give some notations. The examples $x \in \mathcal{X}$ are associated to labels $y \in \{1,-1\}$. The training set is composed of $\ell$ positive examples, so all training labels are $1$. In SVDD, the idea is to find the smallest sphere containing all training data:

\begin{equation}
\label{eq:primalSVDD}
\left\{
  \begin{array}{lll}
    \displaystyle \min_{R,f,\xi} &\displaystyle R + C\sum_{i=1}^{\ell} \xi_i& \\
    \mbox{s.t.} & \displaystyle f(x_i) \leq R + \xi_i & \forall i \in {1\dots \ell} \\
     & \xi_i \geq 0 & \forall i \in {1\dots \ell} \\
  \end{array}
\right.
\end{equation}
where $f(x_i) = k(x_i - a,x_i-a)$, $a$ is the center of the sphere and $R$ is the radius. The kernel function $k(.,.)$ is chosen to be symmetric and semi-definite positive and defines the reproducing kernel Hilbert space ($\mathcal{H}$) containing $f$. 
Note that as discussed in \cite{Chang_arevisit}, if $C>1$, then the problem is equivalent to a hard-margin case. 

The dual is obtained using the Lagrangian technique:
\begin{equation}
\label{eq:lagrangianSVDD}
\begin{array}{ll}
  \displaystyle\underset{(R,f,\xi,d)}{\mathcal{L}} = & \displaystyle R + C\sum_{i=1}^{\ell} \xi_i  + \sum_{i=1}^{\ell} \alpha_i \Bigl(f(x_i)- R - \xi_i \Bigr) \\
  &\displaystyle - \sum_{i=1}^{\ell}\beta_i \xi_i \\
\end{array}
\end{equation}
with $\alpha_i \geq 0 \quad \forall i \in {1\dots\ell}$, $\beta_i \geq 0 \quad \forall i \in{1\dots\ell}$. The KKT conditions are
\begin{equation}
\label{eq:kktSVDD}
  \left\{
    \begin{array}{lcl}
      \displaystyle  \nabla_R \mathcal{L}  = 0 &  \Rightarrow & \displaystyle \sum_{i=1}^{\ell}\alpha_i = 1\\
      \displaystyle  \nabla_{\xi_i} \mathcal{L}  = 0 & \Rightarrow &  C - \alpha_i -\beta_i = 0\\
      \displaystyle  \nabla_f \mathcal{L}  = 0 & \Rightarrow & \displaystyle  k(a,.) = \sum_{i=1}^{\ell} \alpha_i k(x_i,.)  \\
    \end{array}
  \right.
\end{equation}
One can note that using $f(x_i) =  k(x_i - a,x_i-a)$ leads to 
\begin{equation}
\begin{array}{ll}

\displaystyle f(x_i) = & \displaystyle k(x_i,x_i) - 2 \sum_{j=1}^{\ell} \alpha_j k(x_i,x_j)\\ 
& \displaystyle +  \sum_{j=1}^{\ell} \sum_{k=1}^{\ell} \alpha_j \alpha_k k(x_j,x_k)
\end{array}
\end{equation}
The KKT condition are plugged in the Lagrangian which gives the dual problem
\begin{equation}
\label{eq:dualSVDD}
\left\{
  \begin{array}{lll}
    \displaystyle \max_{\alpha} & \displaystyle - \sum_{i=1}^{\ell} \sum_{j=1}^{\ell} \alpha_i \alpha_j k(x_i,x_j) & \displaystyle +  \sum_{i=1}^{\ell} \alpha_i k(x_i,x_i) \\
    \mbox{s.t.} & \displaystyle \sum_{i=1}^{\ell}\alpha_i = 1 &  \\
    & 0 \leq \alpha_i \leq C & \forall i \in {1\dots \ell} \\
  \end{array}
\right.
\end{equation}

\subsection{MultipleKernel SVDD}
In MKL methods, we consider the convex combination of a set of $nk$ kernels such that $k(x_i,x_j) = \sum_{m=1}^{nk} d_m k_m(x_i,x_j)$ with $d_m\geq 0 \quad \forall m \in {1\dots nk}$ and $\sum_{m=1}^{nk}d_m = 1$. Using this multiple kernel, the SVDD primal problem becomes
\begin{equation}
\label{eq:primalMKSVDD}
\left\{
  \begin{array}{lll}
    \displaystyle \min_{R,\{f_m\},\xi,d} &\displaystyle R + C\sum_{i=1}^{\ell} \xi_i & \\
    \mbox{s.t.} & \displaystyle \sum_{m=1}^{nk} d_m f_m(x_i) \leq R + \xi_i & \forall i \in {1\dots \ell} \\
    & \xi_i \geq 0 & \forall i \in {1\dots \ell} \\
     & d_m \geq 0 & \forall m \in {1\dots nk} \\
     & \displaystyle \sum_{m=1}^{nk} d_m=1 &  \\
  \end{array}
\right.
\end{equation}
where $f_m(x_i) = k_m(x_i - a,x_i-a)$, $a$ is the center of the sphere and $R$ is the radius. 
The Lagrangian of system \ref{eq:primalMKSVDD} is
\begin{equation}
\label{eq:lagrangianMKSVDD}
\begin{array}{ll}
  \displaystyle\underset{(R,\{f_m\},\xi,d)}{\mathcal{L}} = & \displaystyle R + C\sum_{i=1}^{\ell} \xi_i - \sum_{i=1}^{\ell}\beta_i \xi_i  \\ & \displaystyle + \sum_{i=1}^{\ell} \alpha_i \Bigl(\sum_{m=1}^{nk} d_m f_m(x_i)- R - \xi_i \Bigr) \\
  &\displaystyle -   \sum_{m=1}^{nk} \gamma_m d_m - \mu (\sum_{m=1}^{nk} d_m - 1) \\

\end{array}
\end{equation}
with $\alpha_i \geq 0 \quad \forall i \in {1\dots\ell}$, $\beta_i \geq 0 \quad \forall i \in{1\dots\ell}$ and $\gamma_m \geq 0 \quad \forall m \in {1\dots nk}$. The KKT conditions are
\begin{equation}
\label{eq:kktMKSVDD}
  \left\{
    \begin{array}{lcl}
      \displaystyle \nabla_R \mathcal{L}  = 0 & \Rightarrow & \displaystyle  \sum_{i=1}^{\ell}\alpha_i = 1\\
      \displaystyle \nabla_{\xi_i} \mathcal{L} = 0 & \Rightarrow & \displaystyle  C - \alpha_i  - \beta_i = 0\\
      \displaystyle \nabla_{f_m} \mathcal{L} = 0  & \Rightarrow & \displaystyle k_m(a,.) = \sum_{i=1}^{\ell} \alpha_i k_m(x_i,.)  \\
      \displaystyle \nabla_{d_m} \mathcal{L} = 0 &\Rightarrow& \displaystyle \sum_{i=1}^{\ell} \alpha_i f_m(x_i) - \gamma_m -\mu = 0\\
    \end{array}
  \right.
\end{equation}
The KKT condition are plugged in the Lagrangian which gives the dual problem
\begin{equation}
\label{eq:dualMKSVDD}
\left\{
  \begin{array}{llll}
    \displaystyle \max_{\alpha,\mu} & \mu && \\
    \mbox{s.t.} & \displaystyle \sum_{i=1}^{\ell}\alpha_i = 1 && \\
    & 0 \leq \alpha_i \leq C & \forall i \in {1\dots \ell} &\\
     &\displaystyle  \sum_{i=1}^{\ell} \alpha_i k_m(x_i,x_i) &-\displaystyle\sum_{i=1}^{\ell} \sum_{j=1}^{\ell} \alpha_i \alpha_j k_m(x_i,x_j) \geq \mu  \\ && \forall m \in {1\dots nk} \\
  \end{array}
\right.
\end{equation}

\subsection{SVDD in SimpleMKL}
The SimpleMKL algorithm is used to solve this problem:
\begin{equation}
\label{eq:solvePrimalMKSVDD1}
\left\{
\begin{array}{ll}
\displaystyle \min_d J(d) &\\ 
\mbox{such that} &\displaystyle \sum_{m=1}^{nk}d_m = 1\\ 
& d_m\geq 0 \quad \forall m \in {1\dots nk}\\
\end{array}
\right.
\end{equation}
where $J(d)=$
\begin{equation}
\label{eq:solvePrimalMKSVDD2}
\left\{
  \begin{array}{lll}
    \displaystyle \min_{R,\{f_m\},\xi} &\displaystyle R + C\sum_{i=1}^{\ell} \xi_i & \\
    \mbox{s.t.} & \displaystyle \sum_{m=1}^{nk} d_m f_m(x_i) \leq R + \xi_i & \forall i \in {1\dots \ell} \\
    & \xi_i \geq 0 & \forall i \in {1\dots \ell} \\
  \end{array}
\right.
\end{equation}

To be able to solve this problem, we need to compute the gradient of $J(d)$. Following the methodology proposed in \cite{rakotomamonjy2008simplemkl}, from system \ref{eq:dualSVDD} we have
\begin{equation}
\label{eq:Jd}
\begin{array}{ll}
J(d) = & \displaystyle - \sum_{i=1}^{\ell} \sum_{j=1}^{\ell} \alpha_i^{\star} \alpha_j^{\star} \sum_{m=1}^{nk} d_m k_m(x_i,x_j) \\
& \displaystyle +  \sum_{i=1}^{\ell} \alpha_i^{\star} \sum_{m=1}^{nk} d_m k_m(x_i,x_i) 
\end{array}
\end{equation}
with $\alpha^{\star}$ maximizes system \ref{eq:dualSVDD} using $k(x_i,x_j) = \sum_{m=1}^{nk} d_m k_m(x_i,x_i)$.

Differentiating $J(d)$ with respect to $d_m$ gives
\begin{equation}
\label{eq:gradJd}
\displaystyle \frac{\partial J}{\partial d_m} =  - \sum_{i=1}^{\ell} \sum_{j=1}^{\ell} \alpha_i^{\star} \alpha_j^{\star} k_m(x_i,x_j) +  \sum_{i=1}^{\ell} \alpha_i^{\star} k_m(x_i,x_i) 
\end{equation} 
which holds for all $m \in {1\dots nk}$

From equations \ref{eq:Jd} and \ref{eq:gradJd}, it is fairly easy to implement SVDD in SimpleMKL, using an SVDD solver to obtain $\alpha^{\star}$.

\subsection{Slim-MK-SVDD}
Multiple kernel SVDD seeks for a combination of kernels that minimizes the radius of the enclosing ball in the kernel space. In practice, it turns out that if Radial Basis Function (rbf) kernels with large bandwidths are available, those will almost always be selected and the frontier around the data in the feature space is loose. In the same way, low degree polynomial kernels will be selected.
This may not be what one is looking for. Hence we propose here to modify the objective function for the optimization of the kernels weights, such that it penalizes loose solutions compared to tighter ones. To do so we observe that tighter solutions rely on more support vectors than loose ones.

\begin{equation}
\label{eq:solvePrimalSMKSVDD1}
\left\{
\begin{array}{ll}
\displaystyle \min_d J(d) - \lambda \mbox{ card}(\alpha) &\\ 
\mbox{such that} &\displaystyle \sum_{m=1}^{nk}d_m = 1\\ 
& d_m\geq 0 \quad \forall m \in {1\dots nk}\\
\end{array}
\right.
\end{equation}
where $J(d)$ is the same as previously, $\lambda$ is a parameter and card$(\alpha)$ is the number of support vectors. 
From the implementation point of view, the SVDD solver can still be used as it is, and the new term only needs to be added at each computation of $J(d)$ and subtracted when computing the (dual-gap based) stopping criteria inside SimpleMKL.

\subsection{Slim-MK-OCSVM}
The {\em slim} extension applied to MK-SVDD is easily applied to MK-OCSVM (multiple kernel one-class SVM) so it was also implemented for the sake of fair comparisons in the experimental part. 

\section{Algorithm evaluation}
In this section we use different one-class algorithms: 
\begin{itemize}
  \item SVDD, 
  \item OCSVM, 
  \item MK-OCSVM ,
  \item Slim-MK-OCSVM, 
  \item MK-SVDD 
  \item Slim-MK-SVDD. 
\end{itemize}
The two first algorithms are standard while the third is described in \cite{rakotomamonjy2008simplemkl}. The others are contributions of this paper.

\subsection{Illustration of the behavior depending on $\lambda$\label{sec:2Dtoy}}
\begin{figure}[ht]
\includegraphics[width=\linewidth]{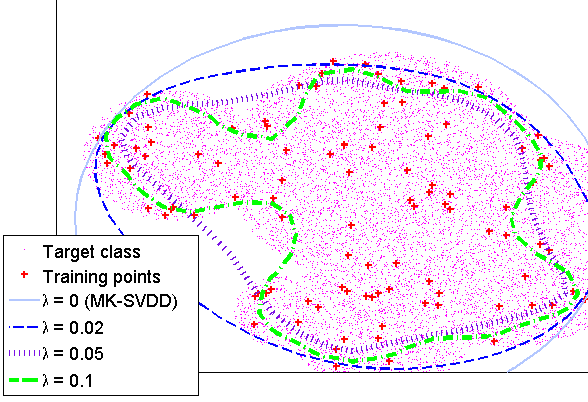}
\caption{\label{fig:evolution} Decision boundaries for Slim-MK-SVDD for increasing values of $\lambda$}
\end{figure}

For this illustration, a 2D target classes is randomly generated. Figure \ref{fig:evolution} shows the decision boundaries given by Slim-MK-SVDD for different values of $\lambda$. We observe that the introduction of this constraint in the problem leads to more fitted results. 

\subsection{Comparison to others one-class kernel methods on 2d problems}
We present here 2 experiments sharing the same setting. We use each of the above mentioned algorithms on 2D targets classes, randomly generated such that there are from 1 to 3 areas. 
All algorithms share the parameter $C$ that varies  in $[0.05, 0.5]$ by steps of $0.05$. For multiple kernel, we use a Gaussian kernel with the following possible bandwidths : $[0.1 ,0.5, 1, 5, 10, 50, 100]$ and a Polynomial kernel of degree 1 to 4. For single kernel methods, we run them for each kernel parameter and keep the best on a validation set (containing only positive examples). 

\begin{figure}[ht]
\includegraphics[width=\linewidth]{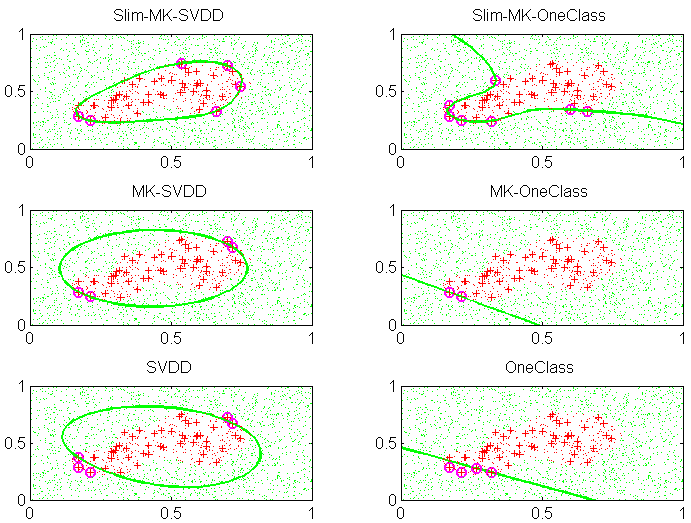}
\caption{\label{fig:toy2d} Visual comparison of performances of different one-class algorithms}
\end{figure}

First we present the decision boundaries for each algorithm on figure \ref{fig:toy2d} for one of the generated problems. 

Second we provide a performance evaluation, computed on average over 25 different target classes. Note that OCSVM-based methods are penalized by the fact that polynomial kernels are used. If removed from the experiment, SVDD-based and OC-SVM based methods are much more comparable. 
\begin{table}
\small
\begin{tabular}{|c|cc|cc|}
\hline
Method & Perf & $\sigma$ & Perf & $\sigma$ \\
 & (rbf) & & (rbf+poly) & \\
\hline
Slim-MK-SVDD & \bf 85.1 & 7.4 & \bf 91.4 & 4.0 \\
Slim-MK-OCSVM & 74.5 & 16.8& 58.0 & 10.7 \\
MK-SVDD & 76.1 & 7.8 & 87.9 & 6.3\\
MK-OCSVM & 33.0 & 18.5 & 40.5 & 9.1\\
SVDD  & 66.1 & 30.4 & 87.0 & 7.1\\
OCSVM  & 60.5 & 33.4& 80.2 & 19.9\\
\hline
\end{tabular}
\caption{\label{tab:toy2d} Average performance and standard deviation on 25 runs}
\end{table}

\subsection{Unsupervised outliers detection \label{sec:unsupervised_outlier}}
In this part we present experiments using a real dataset, namely Breast Cancer from UCI \cite{asuncion2007uci}, modified in order to propose outliers detection task by \cite{amer2013enhancing}. The resulting dataset contains 367 examples described by 30 attributes, with 10 outliers. We also work with Pen-digit local, from the same source, which contains 809 examples described by 16 attributes, with 90 outliers. In the unsupervised case, the complete dataset is used for training, then we check which examples where considered as outliers (i.e. misclassified).

\begin{figure*}[ht]
\centering
\subfloat[\label{fig:outlierBC} Breast Cancer dataset]{
\includegraphics[width=0.47\textwidth]{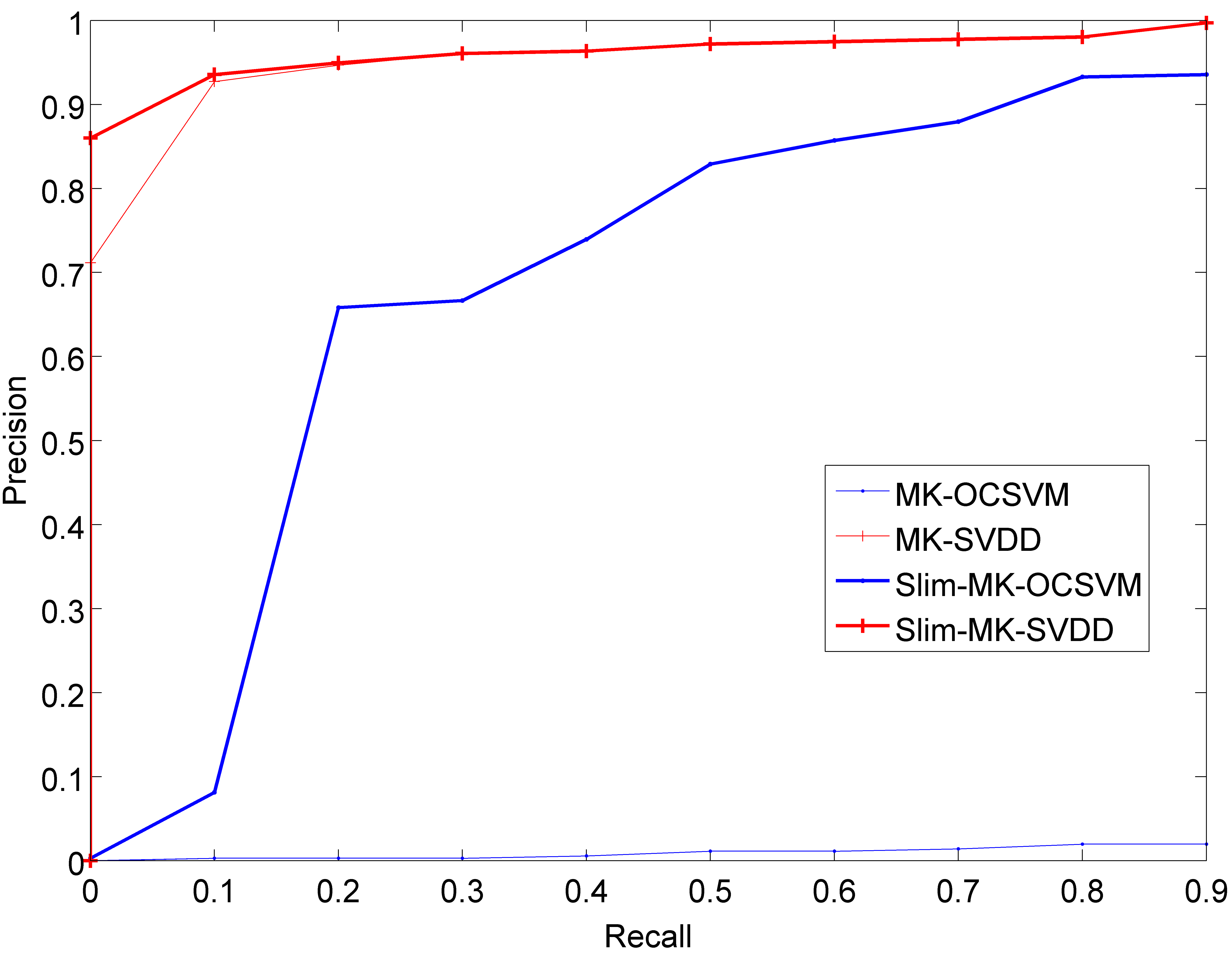}}
\subfloat[\label{fig:outlierPD} Pen Digits dataset]{
\includegraphics[width=0.47\textwidth]{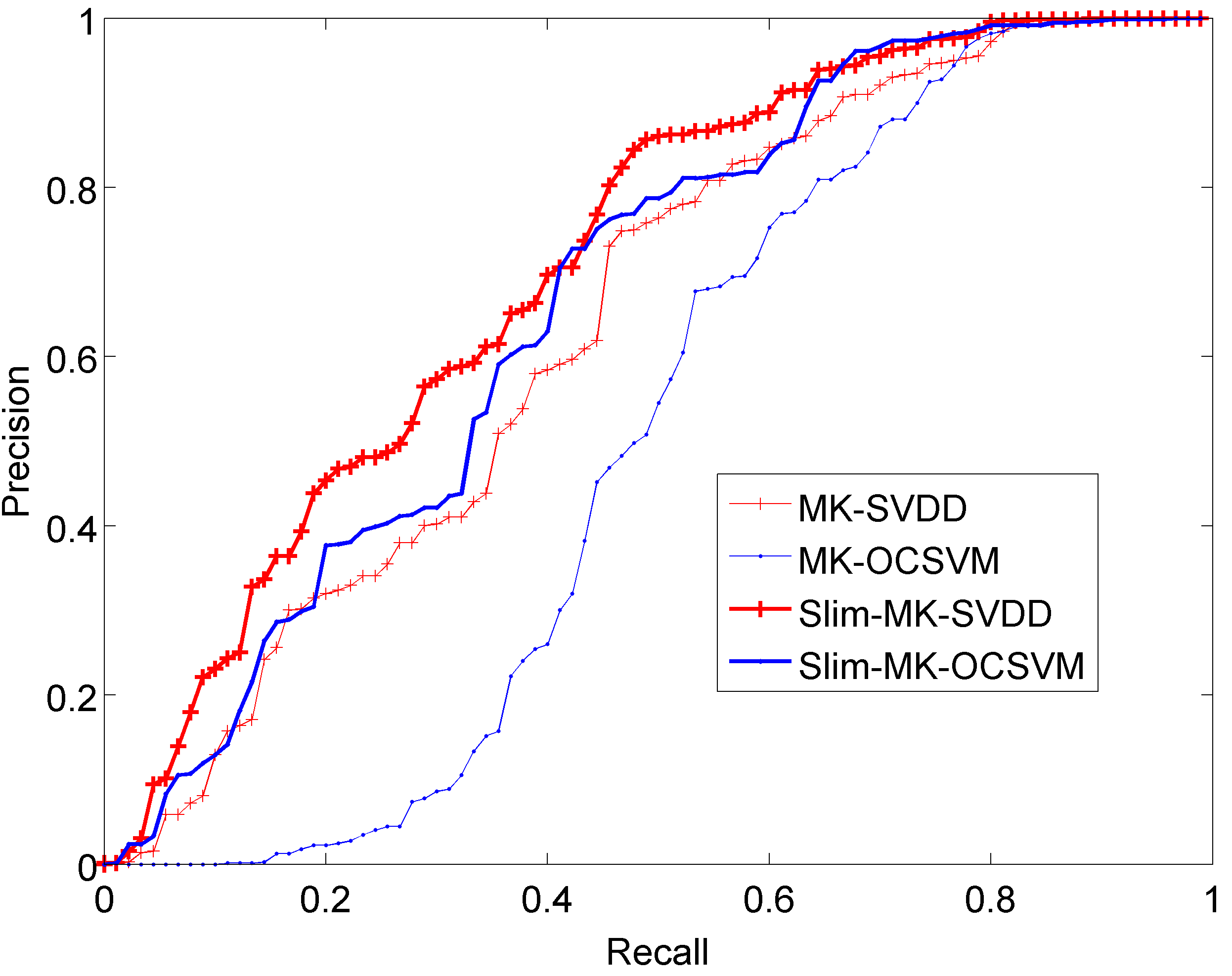}}

\caption{Precision/Recall curves for unsupervised outliers detection, comparing Slim-MK-SVDD and Slim-MK-OCSVM. For each, we present the best ROC curve depending on the $C$ and $\lambda$ parameters.}
\end{figure*}

On figure \ref{fig:outlierBC}, the Precision/Recall curves are plot for Slim-MK-SVDD and Slim-MK-OCSVM on the modified Breast Cancer dataset. For each we present the curve corresponding to the highest AUC (Area Under Curve). Here Slim-MK-SVDD clearly outperforms Slim-MK-OCSVM. Moreover, 9 of the 10 outliers can be discarded with a recall of $0$, that is to say before any false alarm.

Similarly, figure \ref{fig:outlierPD} present the same results for the Pen Digits dataset. This dataset is more difficult. The Slim-MK-SVDD produces the best precision-recall curve, and the {\em Slim} variants improve the performance of both MK-SVDD and MK-OCSVM.

\subsection{Supervised outliers detection \label{sec:supervised_outlier}}
In this section we use the same dataset based on Breast Cancer, but the experiment setting is different since the training is supervised, meaning that only positive examples are used at the training stage, and the resulting models are evaluated on a separated test set, composed of both positive and negative examples.
Moreover, the size of the training set varies in order to observe how many examples are required to be able to obtain a good performance. This performance is measured by the AUC criteria (Area Under Curve). 
For each training set size, training examples are picked randomely among positive ones.
For each training set, $\lambda$ and $C$ are cross-validated. The process is repeated 10 times and results are given on average. 

\begin{figure*}[ht]
\centering
\subfloat[\label{fig:supervisedBC} Breast Cancer dataset]{
\includegraphics[width=0.48\textwidth]{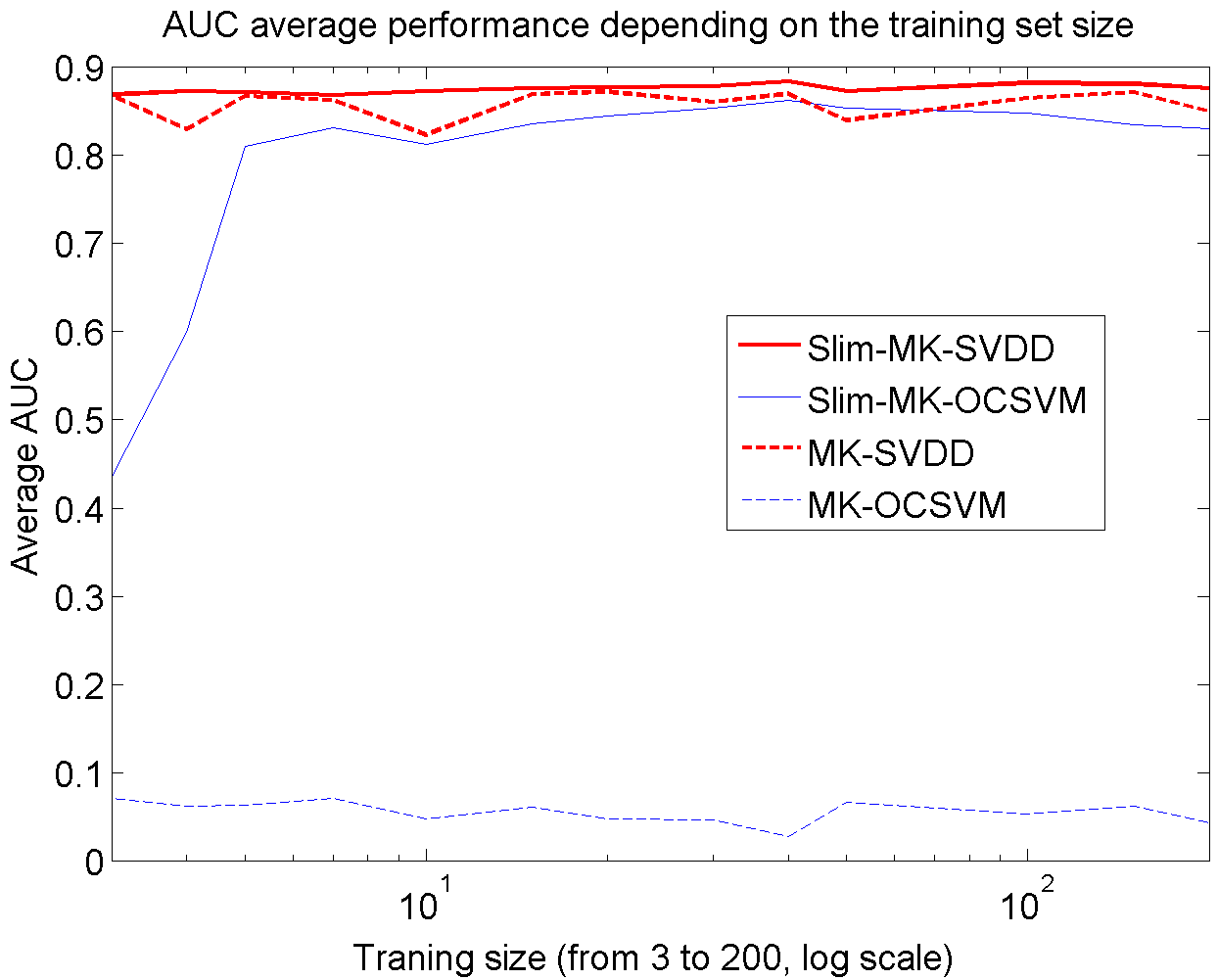}}
\subfloat[\label{fig:supervisedPD} Pen Digits dataset]{
\includegraphics[width=0.47\textwidth]{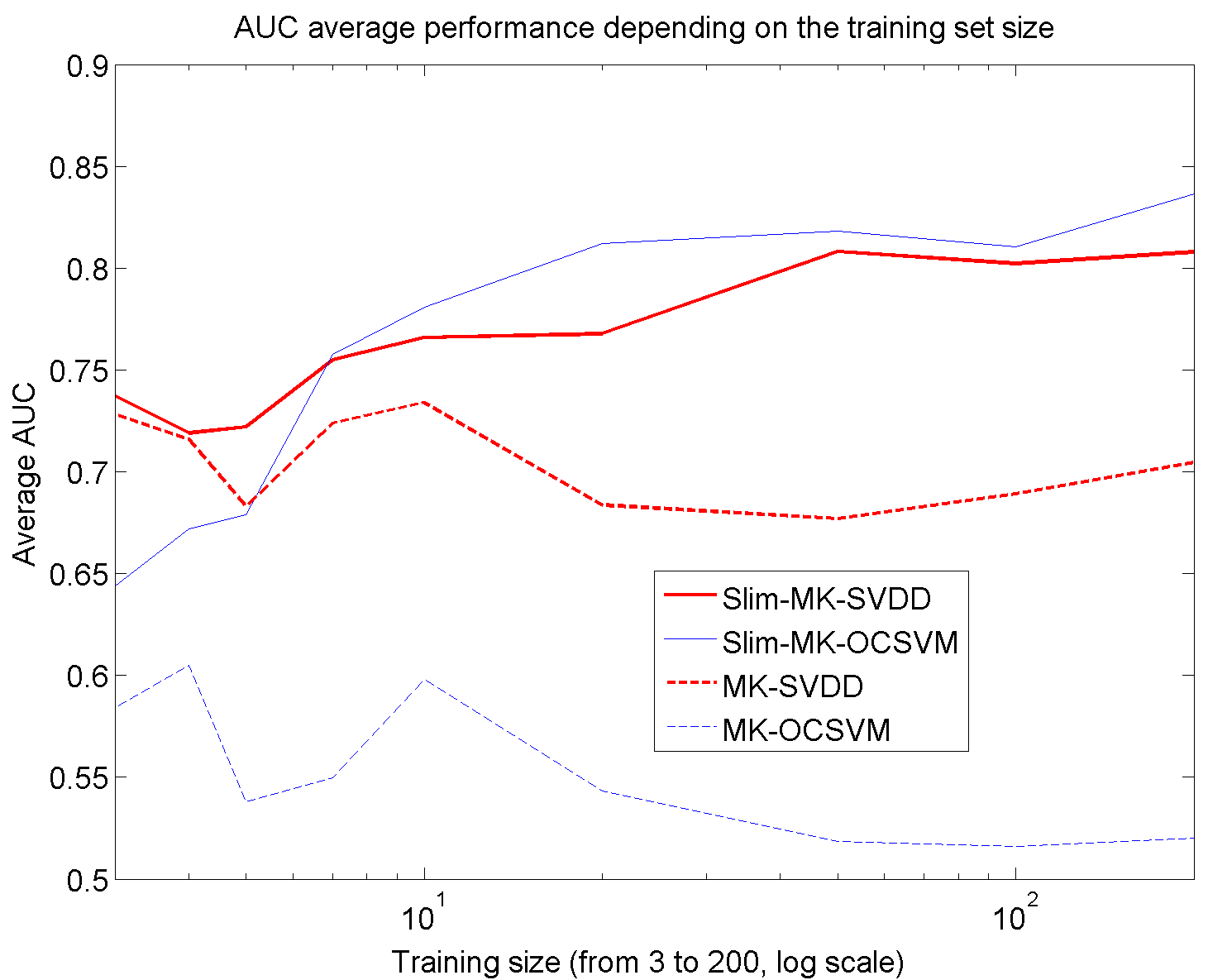}}
\caption{AUC criteria depending on the size of the training set in the supervised case. The traning set size goes from 3 to 200.}
\end{figure*}

On figure \ref{fig:supervisedBC}, the evolution of the AUC depending on the training set size (from 3 to 200) is represented for Slim-MK-SVDD, MK-SVDD, Slim-MK-OCSVM and MK-OCSVM. We observe that for small training set size, the SVDD based methods performs much better than OCSVM based methods. We also observe that MK-OCSVM is completely out of the run, mainly because the linear kernel available in the dictionary of kernels is almost always selected. Added the Slim modification significantly help in this case. 

Similarely, figure \ref{fig:supervisedPD} presents the same results for the Pen Digits dataset.  In this case, we observe that Slim-MK-OCSVM performs better than Slim-MK-SVDD except for very small training sizes. However we observe the same improvement by applying the {\em Slim} variant.

\section{3D-shapes filtering}
In this section we present our motivating application for the development of those algorithms. 

\subsection{Overview of the applicative context}
The 3D-shapes filtering task comes from a large application system dealing with the retrieval of relevant 3D-shapes from a semantic request. Basically, the user is supposed to query for a concept, defined by a (known) class and a (possibly new) adjective or descriptor. For instance, a valid query would be a {\em standing humanoid} or a {\em low chair}. The complete process is described in \cite{aboubacar14}. One step of the process, composed of filters, is meant to select shapes that match the query. The candidate shapes are all generated according to the query class, but it is the filter's task to select those corresponding to the adjective or descriptor. 
In this context, the filter is likely to be trained with very few examples (typically no more than 5) and the proposed shapes are all quite similar (for instance, humanoids in any kind of positions - but always head up).
Moreover, examples going through the filter are proposed to the user, hence it is important that the first result (having the highest output value) corresponds to the query, or that at least the first correct shape appears in the few first proposals.

To evaluate our algorithms outside the whole application but in a similar context, we use a 3D-shape database called Shrec09, described in the next section.

\subsection{Shrec 2009 dataset and graph kernels}
Shrec 2009 is a database of 3D-shapes proposed in \cite{akgul2009shrec}. It contains 200 shapes from 14 labeled classes and 6 unlabeled ones. Its first objective is to compare retrieval methods, which consist in searching among available shapes the closest one to a candidate shape. Classes in this database goes by pairs of similar things (from a structural point of view), like chairs and tables or bikes and motorbikes. A picture of all shapes can be found in \cite{hartveldt2009shrec}.
3D-shapes are in general represented using VRML (Virtual Reality Modeling Language) formats. To manipulate such objects with kernel methods, one has to engineer a special kernel. In \cite{BarraB13}, the authors use Extended Reeb graphs to represent the shapes and then apply graph kernels. The Extended Reeb graphs representation is based on the computation of a scalar function on the initial mesh (see \cite{biasotti2008describing}), the graph coding then exhibiting the invariance properties of this function. 

To conduct our experiments, we took graphs computed from 6 different functions (we refer the reader to \cite{BarraB13} for the description of a functions we used) and computed graph kernels based on bags of paths \cite{suard2007kernel}.

We briefly describe those kernels, considering that for each graph $Gr_i$, we collect a bag of paths $H_i$ (from random walk for instance) of size $|H_i|$. A path $h_m$ stored in a bag of path is composed of $L$ vertices denoted $v$ and $L-1$ edges denoted $e$. We use a kernel function $k_v$ (resp. $k_e$) that compute similarities between the labels on vertices (resp. edges). 
From this, the graph kernel is computed as follows:
\begin{equation}
k(Gr_i,Gr_j) = \frac{1}{|H_i|}\frac{1}{|H_j|} \sum_{m:h_m \in H_i}\sum_{l:h_l \in H_j} k_{d_L}(h_m,h_l)
\end{equation}
where $k_{d_L}$ measures the similarity between two paths of length $L$:
\begin{equation}
k_{d_L}(h_m,h_l) = \exp \Bigl( -\frac{d_L(h_m,h_l)^2}{2\sigma^2}\Bigr)
\label{eq:dl}
\end{equation}
and
\begin{equation}
\begin{array}{ll}
d_L(h_m,h_l) = & k_v(v_m(1),v_l(1)) \\
 & * \prod_{i=2}^{n} \Bigl( k_e(e_m(i-1,i),e_l(i-1,i)) \Bigr.\\
 & * k_v(v_m(i),v_l(i)) \Bigl. \Bigr)
\end{array}
\end{equation}
Note that the similarity between two paths of different length is $0$.

\subsection{Experiment settings}
The evaluation is based on AUC. 
For each run, possible values of $\lambda$ are in $[0, 0.001, 0.01 ,0.1,1]$ and possible values of $C$ are in $[0.01, 0.05, 0.1, 0.2]$.
In each experiment, we apply Slim-MK-SVDD and Slim-MK-OCSVM. When $\lambda=0$, they are identical to MK-SVDD and MK-OCSVM.
Finally, the kernel's dictionary contains 216 kernels from 6 different graphs on each 3D-Shape, depending on the maximum length $L$ of paths in the bag of paths and on the kernel's bandwidths (on edges, vertices and $d_L$ - eq. \ref{eq:dl}). 

\paragraph{Experiment 1.} The first experiment evaluates the ability of our algorithm to differentiate similar classes, when the model is trained on only one class. To do so, we extract in Shrec09 database two similar classes (1 and 2, or 3 and 4, etc.), train on half of the available example from one class (target class), and test on the remaining examples (target and similar). In practice, the training is done on 5 examples. 
Results are given on average, on 10 random sampling of the training set. For each sampling, the hyper-parameters $\lambda$ and $C$ are cross-validated. Results are reported in table \ref{tab:shrek2009} and commented in section \ref{sec:results}.

\paragraph{Experiment 2.} The second experiment evaluates the performance of our algorithm on the complete database. The process is similar to the previous one except that the training set is obtained by using 80\% of the available examples of the target class. As a consequence, only 2 positive examples remains in the test set. 
At each run, the position of the first proposed target class (sorting the output of all test example by descending order) is checked, as well as the position of the first similar class example.
Results, given on average over 10 random split of the target class, are reported in table \ref{tab:shrek2009-all} and commented in section \ref{sec:results}.

\subsection{Results \label{sec:results}}

\paragraph{Experiment 1.} Results provided in table \ref{tab:shrek2009} show for each tested class the best AUC on average, and the corresponding rank 1 performance (Win Ratio). The rank 1 performance corresponds to the proportion of trained filter for which the highest ranked test example is from the target class. We observe that this criteria is not always corresponding to the best AUC. 
From this table we can conclude that:
\begin{itemize}
  \item some pairs of classes are more difficult to distinguish than others
  \item SVDD based methods tends to perform better than OCSVM based methods
  \item OCSVM based methods fails really badly for 6 of the 14 classes (never rank first a target class example in any of the 10 runs)
  \item SVDD based methods seem to be more stable in terms of accuracy
\end{itemize}
Moreover we observed that in general, while the Slim version does not necessarily improve the AUC score or the Win Ratio, it does not degrade the results either (for reasonable values of $\lambda$).

\begin{table}[ht]
\begin{center}
\small
\begin{tabular}{|c|cc|}
  \hline
  Target (Similar) & Best & Win Ratio\\
  Method & AUC &  (rank 1)\\
  \hline
   1 (2) &&\\
   MK-SVDD &  \bf 0.71 &  \bf \em100\% \\
   Slim-MK-OCSVM &  0.56 & \em  80\% \\
  \hline
   2 (1) &&\\
   MK-SVDD &   0.48 & \bf \em 50\% \\
   MK-OCSVM &  \bf 0.51 &  \em 40\% \\
  \hline
   3 (4) &&\\
   MK-SVDD & 0.57  & \bf \em 70\% \\
   MK-OCSVM &  \bf 0.62 &  0\% \\
  \hline
   4 (3) &&\\
   Slim-MK-SVDD & \bf 0.48 & 30\% \\
   MK-OCSVM &  0.43 &  \bf \em 70\% \\
  \hline
   5 (6) &&\\
   MK-SVDD &   0.52 & \em 60\% \\
   MK-OCSVM & \bf 0.80 & \bf \em 100\% \\
  \hline
   6 (5) &&\\
   MK-SVDD &  \bf 0.71 & \bf \em 100\% \\
   MK-OCSVM &  0.40 &  0\% \\
  \hline
   7 (8) &&\\
   MK-SVDD &  \bf 0.73 & \bf \em 90\% \\
   MK-OCSVM &  0.60 &  20\% \\
  \hline
   8 (7) &&\\
   Slim-MK-SVDD &  \bf 0.58  & \em 80\% \\
   MK-OCSVM &  0.51 & \bf \em 100\%  \\
  \hline
   9 (10) &&\\
   Slim-MK-SVDD &  \bf 0.56  & \bf \em 80\% \\
   MK-OCSVM &  0.41 & \em 50\%  \\
  \hline
   10 (9) &&\\
   MK-SVDD &  0.54  & \bf \em 70\% \\
   MK-OCSVM & \bf 0.66  & 0\%  \\
  \hline
   11 (12) &&\\
   MK-SVDD &  0.59  & \em 90\% \\
   MK-OCSVM &  \bf 0.90 & \bf \em 100\%  \\
  \hline
   12 (11) &&\\
   MK-SVDD &  \bf 0.90  & \bf \em 100\% \\
   MK-OCSVM &   0.01 &  0\%  \\
  \hline
   13 (14) &&\\
   Slim-MK-SVDD &  0.53  &  \em 60\% \\
   MK-OCSVM &   \bf 0.57 &  \em 60\%  \\
  \hline
   14 (13) &&\\
   MK-SVDD &   0.46  & \bf 20\% \\
   MK-OCSVM &  \bf 0.48 &  0\%  \\
  \hline

\end{tabular}
\end{center}
\caption{\label{tab:shrek2009} Task: differentiate similar classes. The filter (using either Slim-MK-SVDD or Slim-MK-OCSVM) is trained with a few examples of the target class and is evaluated using both examples from the target class and the similar class. Results are given on average over 10 random splits train/test data. The Win Ratio reports the frequency at which the the example with the highest output value is from the target class. Note that the random Win Ratio is 33\%, results in {\em italic} are better than random. }
\end{table}



\paragraph{Experiment 2.} Results provided in table \ref{tab:shrek2009} show for each tested class, the best AUC on average, and the corresponding average position of the first examples from the target and similar class. 
From those results, we can conclude that:
\begin{itemize}
  \item In most cases, SVDD based methods outperforms OCSVM based methods.  
  \item The Slim variant can improve the performances
  \item It remains a difficult task to filter out the similar class
\end{itemize}

\begin{table}[ht]
\begin{center}
\small
\begin{tabular}{|c|ccc|}
  \hline
  Target (Similar) & Best & Rank of & Rank of\\
  Method & AUC & $1^{st}$ target & $1^{st}$ similar \\
  \hline
   1 (2) &&&\\
   MK-SVDD &  \bf 0.94 & \bf 4.5 & \bf 2\\
   Slim-MK-OCSVM &  0.64 & 55.6 & 34.2\\
  \hline
   2 (1) &&&\\
   MK-SVDD &  \bf 0.80 & \bf 13.5 & \bf 4.8\\
   Slim-MK-OCSVM &  0.66 & 54.9 & 38\\
  \hline
   3 (4) &&&\\
   Slim-MK-SVDD &  \bf 0.945 & \bf2.6 & \bf2.5\\
   MK-OCSVM &  0.28 & 119.2 & 57.5\\
  \hline
   4 (5) &&&\\
   Slim-MK-SVDD &  \bf 0.79 & \bf 15.9 & \bf 1.9\\
   Slim-MK-OCSVM &  0.35 & 98 & 85.1\\
  \hline
   5 (6) &&&\\
   MK-SVDD &   0.63 & \bf 41.2 & \bf 10.1\\
   MK-OCSVM & \bf 0.64 & 52.7 & 74.6 \\
  \hline
   6 (5) &&&\\
   MK-SVDD &   \bf 0.63 & \bf 48.7 & \bf 48.9\\
   Slim-MK-OCSVM &  0.48 & 82.3 & 53.9 \\
  \hline
   7 (8) &&&\\
   MK-SVDD &    0.66 &  19.4 &  7.9\\
   Slim-MK-OCSVM &  \bf 0.88 & \bf 9.2 & \bf 1.8 \\
  \hline
   8 (7) &&&\\
   MK-SVDD &    0.44 &  63.9 &  21.9\\
   MK-OCSVM &  \bf 0.83 & \bf 10.4 & \bf 24 \\
  \hline
   9 (10) &&&\\
   Slim-MK-SVDD &    \bf 0.90 &  \bf 7.5 &  \bf 2.5\\
   MK-OCSVM &   0.39 & 90.1 & 57.7 \\
  \hline
   10 (9) &&&\\
   MK-SVDD &    \bf 0.85 &  \bf 12.1 &  \bf 3.6\\
   Slim-MK-OCSVM &   0.56 & 63.6 & 42.1 \\
  \hline
   11 (12) &&&\\
   Slim-MK-SVDD &    \bf 0.90 &  \bf 4.0 &  \bf 107.3\\
   MK-OCSVM &   0.93 & 8.7 & 128.1 \\
  \hline
   12 (11) &&&\\
   MK-SVDD &    \bf 0.91 &  \bf 10.6 &  82.3\\
   MK-OCSVM &   0.34 & 114.6 & \bf 2.4 \\
  \hline
   13 (14) &&&\\
   Slim-MK-SVDD &     \bf 0.69 &  \bf 29.7 &  \bf 16.2\\
   MK-OCSVM &    0.48 & 68.7 & 40.9 \\
  \hline
   14 (13) &&&\\
   MK-SVDD &    \bf 0.73 &  \bf 18.2 &  \bf 4.8\\
   MK-OCSVM &   0.44 & 86.3 & 17.4 \\
  \hline

\end{tabular}
\end{center}
\caption{\label{tab:shrek2009-all} Task: differentiate a given class from all others. The filter (using either Slim-MK-SVDD or Slim-MK-OCSVM) is trained with 8 examples of the target class and is evaluated using all remaining examples (all other classes and the 2 remaining positive examples). Results are given on average over 10 random selection of the training examples. When the best average result is obtained for $\lambda = 0$ (or when $\lambda=0$ performs as well as $\lambda>0$), the method is noted without {\em Slim}.}
\end{table}

\section{Discussion and conclusion}
From an applicative problem, consisting in filtering 3D-shapes based on a few examples of valid shapes, we explored the idea of using the SVDD algorithm within the SimpleMKL method. The main reasons for this was the large amount of available kernels to be compared and the fact that SVDD is known to be more flexible than OCSVM, when considering kernel shapes that can be used.

Plugging SVDD in SimpleMKL is almost straight forward but due to the objective function of the MKL (that seeks for the kernel combination corresponding to the "simplest" decision boundary), the proposed solutions are not always well adapted for filtering. Indeed, we may need a boundary that is able to exclude parts of the feature space that are not covered by examples. 
We propose a derivation of MK-SVDD by adding an "over-fitting" term to the objective function of SimpleMKL, consisting in penalizing the solutions based on fewer support vectors. While this approach is not fully explored on a theoretical point of view, the experimental results show that this method can help one-class algorithms such as SVDD or OCSVM to improve their performances in the context of multiple kernel training.

Apart form our application, we believe that MK-SVDD and Slim-MK-SVDD can be successfully applied in other contexts, such as outliers detection. In sections \ref{sec:unsupervised_outlier} and \ref{sec:supervised_outlier}, we illustrate the ability of the presented algorithm to detect outliers in the supervised case (only positive examples are used for training) and unsupervised case (some outliers are contained in the training dataset). Of course those applications require some more extended studies and this is a path we want to follow in the continuation of our researches, together with a deeper study of the theoretical aspects.

\subsection*{Acknowledgment}
The authors wish to thank Silvia Biasotti for the computation of the Extended Reeb Graphs and Vincent Barra for helpful comments and discussions.

\clearpage

\label{sec:la-bibliographie}

\bibliography{biblio}

\end{document}